\documentclass[preprint,5p,times, authoryear, sort, comma, singleblind, nopreprintline]{elsarticle}

\usepackage[utf8]{inputenc}
\usepackage{amsmath,amsfonts,amssymb, mathtools, blindtext}
\usepackage{graphicx}
\usepackage{subfig}
\usepackage{caption}
\usepackage{float} 
\usepackage{subcaption}

\usepackage{graphicx}
\usepackage{multirow}
\usepackage{booktabs} 

\usepackage{natbib}
\usepackage{graphicx}
\usepackage[table,xcdraw]{xcolor}
\usepackage{subfig}
\usepackage{graphicx}

\usepackage{tabularx}
\usepackage{amssymb}
\usepackage{float}
\usepackage{xcolor}
\usepackage{multirow}

\usepackage{xcolor}
\usepackage{hyperref}

\begin{document}

\begin{frontmatter}
\title{Leveraging Unsupervised Learning for Cost-Effective Visual Anomaly Detection}

\author[inst1]{Yunbo Long\corref{cor1}}
\author[inst1]{Zhengyang Ling}
\author[inst1]{Sam Brook}
\author[inst1]{Duncan McFarlane}
\author[inst1]{Alexandra Brintrup}
\affiliation[inst1]{organization={Department of Engineering, University of Cambridge},country={United Kingdom}}

\cortext[cor1]{Corresponding author: yl892@cam.ac.uk}

\begin{abstract}
Traditional machine learning-based visual inspection systems require extensive data collection and repetitive model training to improve accuracy. These systems typically require expensive camera, computing equipment and significant machine learning expertise, which can substantially burden small and medium-sized enterprises. This study explores leveraging unsupervised learning methods with pre-trained models and low-cost hardware to create a cost-effective visual anomaly detection system. The research aims to develop a low-cost visual anomaly detection solution that uses minimal data for model training while maintaining generalizability and scalability. The system utilises unsupervised learning models from Anomalib and is deployed on affordable Raspberry Pi hardware through openVINO. The results show that this cost-effective system can complete anomaly defection training and inference on a Raspberry Pi in just 90 seconds using only 10 normal product images, achieving an F1 macro score exceeding 0.95. While the system is slightly sensitive to environmental changes like lighting, product positioning, or background, it remains a swift and economical method for factory automation inspection for small and medium-sized manufacturers. The code is available at
\href{https://github.com/Yunbo-max/Cost-Effective-Visual-Anomaly-Detection-using-Unsupervised-Learning}
{\textcolor{red}{https://github.com/Yunbo-max/Cost-Effective-Visual-Anomaly-Detection-using-Unsupervised-Learning}}.
\end{abstract}


\end{frontmatter}

\section{Introduction} \label{sec: introduction}

In modern industry, anomaly detection typically occurs during the final product quality inspection phase. When identifying defects, the product's appearance is often one of the first things checked; any visible flaws can lead to the product being discarded, re-manufactured or undergoing further testing. Traditionally, visual anomaly detection has been handled manually, especially for small and medium-sized enterprises \citep{liu2024deep}. However, humans are prone to errors, and the quality of checks can vary from worker to worker \citep{apostolopoulos2023industrial}. In addition, the labour costs associated with manual inspection can be a large part of the cost for SMEs \citep{jha2023deep}. Consequently, many automated visual anomaly detection systems have been introduced into manufacturing. Early computer vision systems that used YOLO (You Only Look Once) for image segmentation recognised defects by segmenting entire images into regions and identifying objects within those regions, rather than segmenting individual pixel points \citep{hussain2023yolo}. In recent years, deep learning-based image recognition systems have used large datasets to learn the characters of both normal and abnormal products; these have provided significant advantages in product defect detection and have been successful in industrial applications \citep{rippel2023anomaly}.

Deep learning models have demonstrated impressive image recognition and anomaly detection classification capabilities. However, its application in industrial anomaly detection is heavily based on large-scale data sets of labelled or classified images of products \citep{zipfel2023anomaly}. Gathering and annotating this data for training deep learning models requires significant human labour and time. Additionally, due to the diversity of product defects in industrial production, there is a need for continual collection and annotation of different defect images to update the detection models \citep{jiang2022survey}. When manufacturing companies produce different products, deep learning models struggle to transfer to new products and types of defects. This requires the collection of new data, which is a challenge for small and medium-sized enterprises (SMEs) \citep{jha2023deep}. Consequently, SMEs often lack the capacity to collect and maintain sufficient image data to effectively train deep learning models \citep{xie2024iad}. Typical industrial deep learning detection systems have high requirements for GPU computing power and camera resolution, further hindering the adoption of intelligent detection solutions by SMEs \citep{tao2022deep}. These factors collectively contribute to the reluctance of SMEs to apply deep learning models for practical anomaly detection problems.

The primary objective of this research is to develop a low-cost, effective machine learning-based visual anomaly detection system tailored for SMEs. Pre-trained models minimise the need for extensive data collection and reduce the training required. The approach involves fine-tuning and comparing existing models that have been implemented on Raspberry Pis. The system has been deployed in lab tests with a gearbox product. The key contributions are shown below.

\begin{enumerate}[I.]
\item \textbf{Creation of a low-cost System:} Development of a visual anomaly detection system using a Raspberry Pi 4B, showcasing an affordable method for detecting defects in gearbox parts.
\item \textbf{Evaluation of Multiple Unsupervised Learning Algorithms:} Application of different pre-trained models for anomaly detection.
\item \textbf{Performance Validation:} Lab tests with a product testing performance with different defects and conditions. 
\end{enumerate}

Section 2 introduces and reviews the algorithms and related work. Section 3 details the study's methodology, including the workflow and describes the experimental setup. Section 4 reviews the model and system performance and the discusses the results setup final section summarizes the findings.

\section{Related work on Anomaly Detection Algorithms}
\subsection{Background}
Anomalib is a deep learning library designed for anomaly detection, offering a modular framework for developing and deploying advanced models \citep{akcay2022anomalib, christie2023enhancing}. It leverages pre-trained convolutional neural networks (CNNs) for feature extraction from datasets like ImageNet, enhancing anomaly detection and localization tasks \citep{russakovsky2015imagenet, bergmann2020uninformed}. Algorithms within Anomalib, such as PaDiM and PatchCore, benefit from these pre-trained models to improve efficiency efficiency and performance \citep{kim2023rag, santos2023optimizing}. Conversely, algorithms like CFlow-AD and Fastflow learn feature representations and data distributions from scratch \citep{kim2023sanflow, zheng2022benchmarking}.

Despite the advancements in anomaly detection, there is limited research on deploying these algorithms with pre-trained models on low-cost hardware such as a Raspberry Pi. For instance, \cite{ziegler2023applications} explored using methods like PatchCore for anomaly detection in manufacturing and suggested that the Raspberry Pi could serve as a processor for such tasks in resource-constrained environments; however, they did not investigate this. Similarly, \cite{hattori2023defect} applied PatchCore to detect defects in food items, such as apples, but did not explore its deployment on low-cost hardware. Moreover, managing uncertainty is essential for ensuring system trustworthiness. \cite{yong2022bayesian} presents Bayesian autoencoders (BAEs) to address this in unsupervised anomaly detection, but there is a lack of research on deploying reliable visual anomaly detection systems on low-cost hardware.

\subsection{PaDiM}
PaDiM (Patch-wise Anomaly Detection via Density Modeling) is used as a method to detect image anomalies and localization by modelling the density of image patches \citep{defard2021padim}. Specifically, it leverages a pre-trained CNN to extract features from image patches. The CNN helps convert raw image data into a more useful form for detecting anomalies. PaDiM uses multivariate Gaussian distributions to model the distribution of these normal image patches \citep{chew2022anomaly}. This probabilistic approach helps determine how likely a given patch is to be part of the normal class \citep{zipfel2023anomaly}. For example, for a production line of electronic components, PaDiM helps detect soldering defects. The pre-trained CNN extracts features from the solder joints, and the Gaussian distributions model helps identify faulty components.

\subsection{PatchCore}

PatchCore is an anomaly detection algorithm that identifies anomalies by comparing blocks of features in an image with a set of “normal” blocks of features stored in memory \citep{roth2022towards}. It uses feature extractors to compute descriptors for image blocks and then performs a k-nearest-neighbour search to detect anomalies based on similarity \citep{roth2022towards}. PatchCore uses pre-trained feature extractors and benefits from optimizing these extractors to improve performance \citep{santos2023optimizing}. Recent advances have shown models with better classification performance on large datasets \citep{jiang2024fr}.

\subsubsection{Other Methods}

CFlow-AD (Conditional Flow-based Anomaly Detection) uses normalizing flows to model the distribution of normal data and detect anomalies based on deviations from this distribution, capturing complex patterns without relying on pre-trained models \citep{gudovskiy2022cflow}. FastFlow is a fast and efficient flow-based method for anomaly detection, optimizing normalising flow models for speed and scalability, and is designed for real-time applications without pre-trained models \citep{yu2021fastflow}.

\subsection{Model deployment}
For deploying these algorithms, the OpenVINO toolkit (Open
Visual Inference and Neural Network Optimization), introduced by \cite{gorbachev2019openvino}, plays a crucial role. OpenVINO optimizes the deployment and inference of deep learning models by reducing their size while maintaining accuracy. This optimization is essential for deploying models on edge devices with constrained computational resources. Moreover, Anomalib supports OpenVINO for model optimisation and quantisation, further enhancing the feasibility of deploying models on optimisation by reducing computational costs.

\section{Methodology}

\subsection{Experiment Settings}
Gears and gear casings are placed on a workstation table for
assembly; before they are assembled, they are checked using
the visual anomaly detection system A camera is positioned above the tray to capture images. The system's role is to verify whether the parts are normal or abnormal before the robots begin the assembly process. 

\begin{figure*}[htbp!] 
    \centering
    \includegraphics[width=0.9\textwidth]{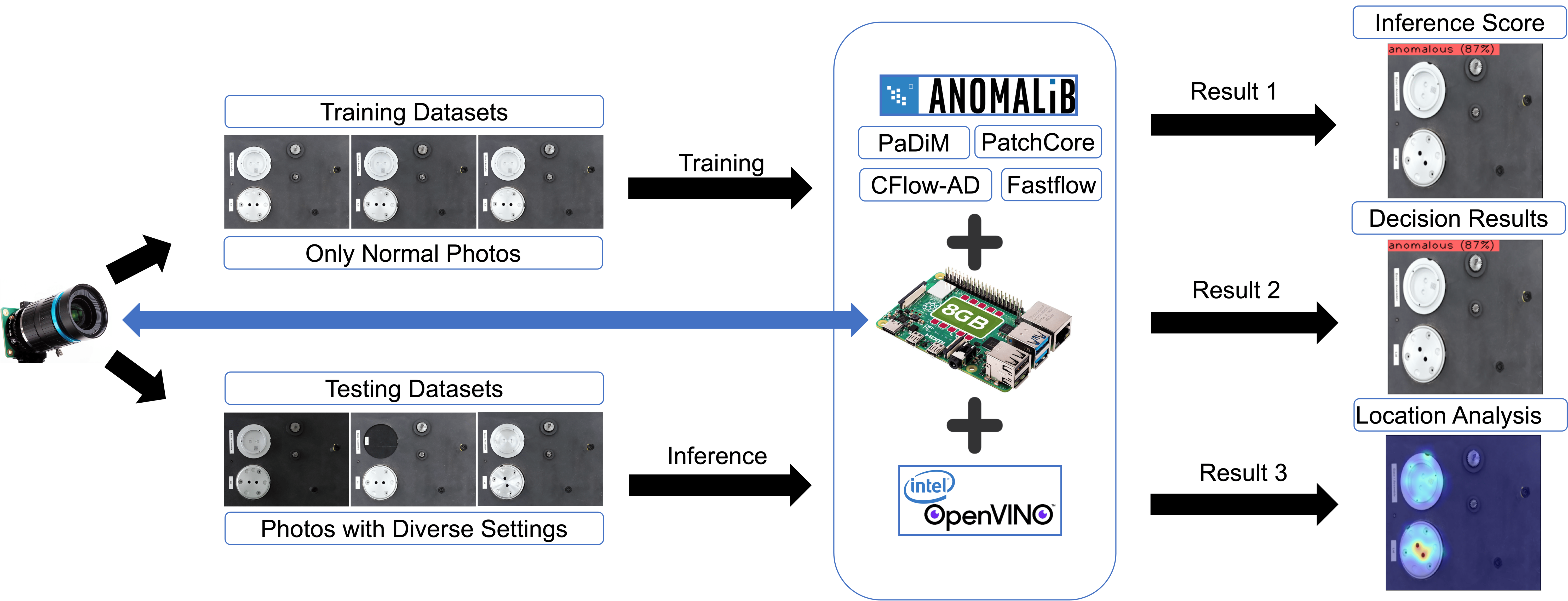}
    \caption{System Workflow}
    \label{fig:workflow}
\end{figure*}

The system workflow shown in Figure \ref{fig:workflow} describes the overall process followed in the work. Initially, low-cost hardware is used to capture 20 images of normal products, see Figure \ref{fig:defects} (a), which are used to train anomaly detection models. Models tested, which are outlined in the previous section, include PaDiM, PatchCore, CFlow-AD, and Fastflow. These are deployed using the ANOMALiB library within the OpenVINO environment. After training, the model is saved and applied to images captured from the lab set-up for anomaly detection. During inference, the system generates a confidence score to assess whether parts are normal or anomalous. Anomalies are highlighted in the images with heat maps if detected. 

\subsection{Hardware}
The Raspberry Pi 4B was used in all tests. \citep{mounir2020performance}  compared the execution time performance of basic image processing algorithms on various platforms. It highlighted that the Raspberry Pi can run lightweight deep learning models efficiently and is comparable to entry-level x86 personal computers (PCs). Raspberry Pis has been successfully employed in various low-cost applications, including face recognition \citep{durr2015deep}, unmanned aerial vehicle (UAV) systems \citep{qurishee2019low}, and bump detection \citep{dewangan2020deep}.  Compared to other microcomputers, it offers greater cost to performance \citep{suzen2020benchmark}, making it an ideal choice for developing low-cost systems. An RPI-6MM LENS camera captures images of parts for training and testing.

\subsection{Data Collection}

\begin{figure*}[htbp!] 
    \centering
    \includegraphics[width=0.9\textwidth]{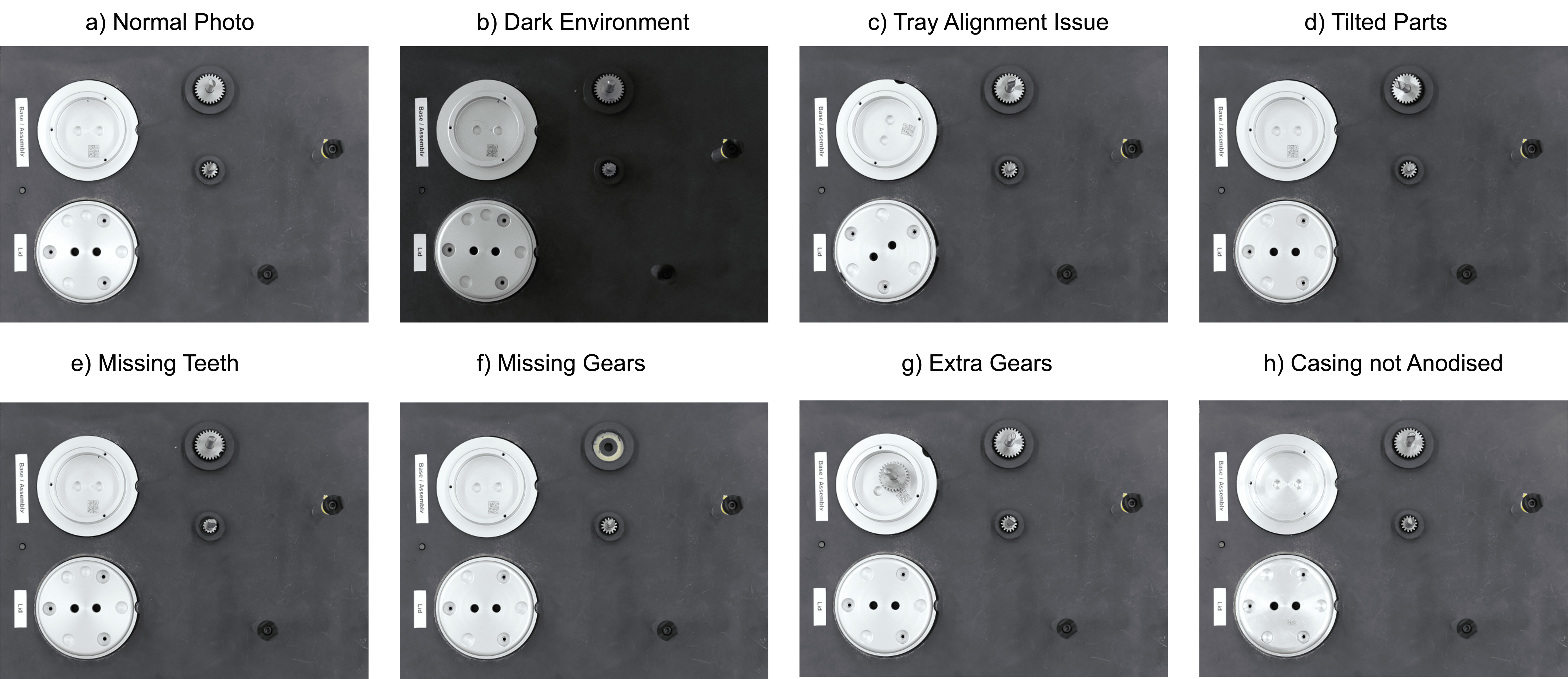}
    \caption{Images of different setup conditions (b, c and d) and different product conditions (e, f, g, and h).}
    \label{fig:defects}
\end{figure*}

Data collection encompasses a range of conditions, including different types of both product and setup conditions. Two conditions are changed to evaluate the model's anomaly detection performance:
\begin{enumerate}[I.]
\item Product condition anomalies - This setup relates to the product condition. Four different types are tested: 1) missing gear teeth from the main gear (\label{fig:defects}e), 2) missing gear part (\label{fig:defects}f), 3) extra gear added to the tray (\label{fig:defects}g), and 4) a casing not anodised (\label{fig:defects}h). Each of these anomalies is shown in pictures in Figure \ref{fig:defects}. These are common problems that need to be picked up by the detection system because they will cause problems with the product's performance or assembly issues. 
\item  Setup condition anomalies- These are changes in the environment which do not represent a fault with the product. These include a change in lighting (\label{fig:defects}b), a movement or misalignment of the tray of parts (\label{fig:defects}c) a rotation or tilting of the parts away from their set position (\label{fig:defects}d). 
\end{enumerate}

\subsection{Anomaly Detection Experiments Setting}

In this experiment, PatchCore, a state-of-the-art visual anomaly detection method on MVTec industrial datasets \citep{santos2023optimizing}, was employed on a Raspberry Pi with with 8GB RAM. For each product and setup condition, the same 20 normal photos were employed. Out of these, 15 photos were randomly chosen for training. The remaining 5 normal photos and 5 diverse photos for the specific conditions were used for testing. It is important to note that the training dataset consists only of photos with correct product and setup conditions. The entire process of training and testing for each setting was performed, and AUROC and F1 scores were recorded to evaluate the model’s performance. 

Beyond the final scores, understanding how the model infers anomalies from unseen photos and processes these photos to identify the anomaly area is crucial for model explanation. Therefore, example figures are provided of the heatmaps created, highlighting where anomalies are picked up. In manufacturing anomaly detection, a confidence score is crucial for assessing the reliability of classification results for each photo. This score indicates the algorithm's certainty about whether a photo is normal or anomalous. Generally, a confidence score above 50\% is needed to outperform random guessing, while a score above 80\% is required for a convincing decision. This is critical because real production environments often introduce noise that can affect model accuracy. The probabilistic score helps workers and managers make informed decisions regarding noisy photos. To thoroughly evaluate model performance, we present detailed model inference results, including confidence scores and defect localization, in the next section.

\subsection{Ablation Study}

To validate the robustness and generalization capability of our anomaly detection system, we conducted ablation studies to evaluate its performance across various hardware setups and when processing different datasets. For this purpose, we redeployed the system in a different gearbox installation environment and collected new sets of normal photos for the components from different production batches. Following the same experimental steps as before, we also gathered abnormal photos corresponding to these new products and setup conditions. The normal photos were then divided into training and testing datasets in a 2:1 ratio. Then, the model was trained exclusively on normal photos, while testing utilized both normal and all abnormal photos. This study aims to examine how varying dataset sizes (20, 40, and 80 images, with 25\% being normal) impact model performance on diverse hardware setups.

Given the limited capacity of the low-cost system, it is essential to record both training and inference times on different devices. A comparative study was conducted between a general personal computer running Microsoft Windows and the Raspberry Pi-based low-cost system. This comparison helps determine if the cost-effective system can achieve performance levels similar to those of a more powerful general-purpose computer.

\subsection{Performance Metrics}

To evaluate the model's performance, we use AUROC and F1 Macro metrics for the tests. In our detection experiment, results are classified as positive (anomaly detected) or negative (normal function). positives are correctly identified anomalies, while false positives are incorrectly predicted anomalies. These results are summarized in Table \ref{tab:confusion_matrix}.

\begin{table}[h!]
    \centering
    \begin{tabular}{|c|c|c|}
        \hline
        & \textbf{Predicted Positive} & \textbf{Predicted Negative} \\
        \hline
        \textbf{Actual Positive} & True Positive (TP) & False Negative (FN) \\
        \hline
        \textbf{Actual Negative} & False Positive (FP) & True Negative (TN) \\
        \hline
    \end{tabular}
    \caption{Confusion Matrix}
    \label{tab:confusion_matrix}
\end{table}

\subsubsection{AUROC}
The AUROC (Area Under the Receiver Operating Characteristic Curve) is a metric used to quantify the model's ability to distinguish between classes. It is defined as the area under the ROC curve, which plots the True Positive Rate (TPR) against the False Positive Rate (FPR) across various threshold settings. Mathematically, AUROC can be expressed as:
\begin{equation}
\text{AUROC} = \int_{0}^{1} \text{TPR}(x) \, d\text{FPR}(x)
\end{equation}

where:
\begin{equation}
\text{TPR} = \frac{\text{TP}}{\text{TP} + \text{FN}}
\end{equation}

\begin{equation}
\text{FPR} = \frac{\text{FP}}{\text{FP} + \text{TN}}
\end{equation}

The ROC curve plots TPR versus FPR, and the AUROC represents the probability that the model will rank a randomly chosen positive instance higher than a randomly chosen negative instance.

\subsubsection{F1 Score}

The F1 score macro balances precision and recall in multi-class classification problems, making it suitable for anomaly detection.

Recall rate and precision are defined as

\begin{equation}
\text{Recall Rate} = \frac{\text{TP}}{\text{TP} + \text{FN}}
\end{equation}

\begin{equation}
\text{Precision} = \frac{\text{TP}}{\text{TP} + \text{FP}}
\end{equation}

The macro F1 score is:

\begin{equation}
\text{F1 Macro} = \frac{1}{C} \sum_{i=1}^{C} \text{F1}_i
\end{equation}

where \(C\) is the number of classes, and \(\text{F1}_i\) is the F1 score for the \(i\)-th class:

\begin{equation}
\text{F1}_i = 2 \times \frac{\text{Precision}_i \times \text{Recall}_i}{\text{Precision}_i + \text{Recall}_i}
\end{equation}

Here, \(\text{Precision}_i\) and \(\text{Recall}_i\) are the precision and recall for the \(i\)-th class.

\section{Results and Discussion}

\begin{table}[H]
\caption{F1 macro and AUROC Test Scores Across Various Conditions and Product Combinations on Raspberry Pi}
\label{tab:F1 ande AUROC}
\resizebox{\columnwidth}{!}{%
\begin{tabular}{|llcccc|}
\hline
\multicolumn{6}{|c|}{\textbf{\begin{tabular}[c]{@{}c@{}}Test Results for Anomaly Detection Model Under Different Conditions\\ (F1 macro and AUROC Score)\end{tabular}}} \\ \hline
\multicolumn{2}{|l|}{\multirow{2}{*}{}} &
  \multicolumn{4}{c|}{\textbf{Setup Condition}} \\ \cline{3-6} 
\multicolumn{2}{|l|}{} &
  \multicolumn{1}{l|}{\textbf{No Change}} &
  \multicolumn{1}{l|}{\textbf{Darker Environment}} &
  \multicolumn{1}{l|}{\textbf{Tray not Aligned}} &
  \multicolumn{1}{l|}{\textbf{Parts Tilt}} \\ \hline
\multicolumn{1}{|l|}{\multirow{5}{*}{\textbf{\begin{tabular}[c]{@{}l@{}}Product   \\ \\ Condition\end{tabular}}}} &
  \multicolumn{1}{l|}{\textbf{Normal Products}} &
  \multicolumn{1}{c|}{1.0 (1.0)} &
  \multicolumn{1}{c|}{1.0 (1.0)} &
  \multicolumn{1}{c|}{1.0 (1.0)} &
  1.0 (1.0) \\ \cline{2-6} 
\multicolumn{1}{|l|}{} &
  \multicolumn{1}{l|}{\textbf{Gear Damage}} &
  \multicolumn{1}{c|}{1.0 (1.0)} &
  \multicolumn{1}{c|}{1.0 (1.0)} &
  \multicolumn{1}{c|}{1.0 (1.0)} &
  1.0 (1.0) \\ \cline{2-6} 
\multicolumn{1}{|l|}{} &
  \multicolumn{1}{l|}{\textbf{\begin{tabular}[c]{@{}l@{}}Casing not Annodised \\ (shiny metal)\end{tabular}}} &
  \multicolumn{1}{c|}{1.0 (1.0)} &
  \multicolumn{1}{c|}{1.0 (1.0)} &
  \multicolumn{1}{c|}{1.0 (1.0)} &
  1.0 (1.0) \\ \cline{2-6} 
\multicolumn{1}{|l|}{} &
  \multicolumn{1}{l|}{\textbf{Missing Gears}} &
  \multicolumn{1}{c|}{1.0 (1.0)} &
  \multicolumn{1}{c|}{1.0 (1.0)} &
  \multicolumn{1}{c|}{1.0 (1.0)} &
  1.0 (1.0) \\ \cline{2-6} 
\multicolumn{1}{|l|}{} &
  \multicolumn{1}{l|}{\textbf{Extra Gears}} &
  \multicolumn{1}{c|}{1.0 (1.0)} &
  \multicolumn{1}{c|}{1.0 (1.0)} &
  \multicolumn{1}{c|}{1.0 (1.0)} &
  1.0 (1.0) \\ \hline
\end{tabular}%
}
\end{table}

\begin{table}[H]
\caption{Number of Anomalies Detected across Various Conditions and Product Combinations on Raspberry Pi}
\label{tab:Number of Anomalies}
\resizebox{\columnwidth}{!}{%
\begin{tabular}{|llcccc|}
\hline
\multicolumn{6}{|c|}{\textbf{Number of Anomalies Detected Under Different Conditions (10 in total)}} \\ \hline
\multicolumn{2}{|l|}{\multirow{2}{*}{}} &
  \multicolumn{4}{c|}{\textbf{Setup Condition}} \\ \cline{3-6} 
\multicolumn{2}{|l|}{} &
  \multicolumn{1}{l|}{\textbf{No Change}} &
  \multicolumn{1}{l|}{\textbf{Darker Environment}} &
  \multicolumn{1}{l|}{\textbf{Tray not Aligned}} &
  \multicolumn{1}{l|}{\textbf{Parts Tilt}} \\ \hline
\multicolumn{1}{|l|}{\multirow{5}{*}{\textbf{\begin{tabular}[c]{@{}l@{}}Product   \\ \\ Condition\end{tabular}}}} &
  \multicolumn{1}{l|}{\textbf{Normal Products}} &
  \multicolumn{1}{c|}{0} &
  \multicolumn{1}{c|}{0} &
  \multicolumn{1}{c|}{5} &
  5 \\ \cline{2-6} 
\multicolumn{1}{|l|}{} &
  \multicolumn{1}{l|}{\textbf{Gear Damage}} &
  \multicolumn{1}{c|}{5} &
  \multicolumn{1}{c|}{5} &
  \multicolumn{1}{c|}{5} &
  5 \\ \cline{2-6} 
\multicolumn{1}{|l|}{} &
  \multicolumn{1}{l|}{\textbf{\begin{tabular}[c]{@{}l@{}}Casing not Annodised \\ (shiny metal)\end{tabular}}} &
  \multicolumn{1}{c|}{5} &
  \multicolumn{1}{c|}{5} &
  \multicolumn{1}{c|}{5} &
  5 \\ \cline{2-6} 
\multicolumn{1}{|l|}{} &
  \multicolumn{1}{l|}{\textbf{Missing Gears}} &
  \multicolumn{1}{c|}{5} &
  \multicolumn{1}{c|}{5} &
  \multicolumn{1}{c|}{5} &
  5 \\ \cline{2-6} 
\multicolumn{1}{|l|}{} &
  \multicolumn{1}{l|}{\textbf{Extra Gears}} &
  \multicolumn{1}{c|}{5} &
  \multicolumn{1}{c|}{5} &
  \multicolumn{1}{c|}{5} &
  5 \\ \hline
\end{tabular}%
}
\end{table}

\begin{figure}[h!]
        \centering
        \includegraphics[width=0.45\textwidth]{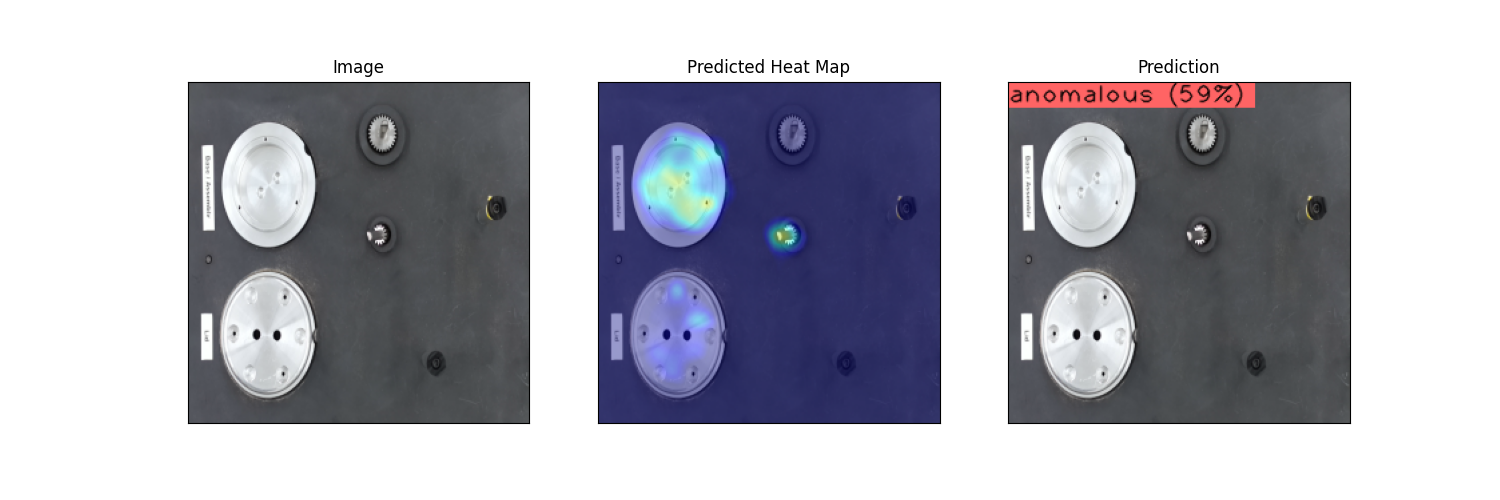}
        \caption{Aonomaly Detection Results for Non-Anodized Tilt Parts}
        \label{fig:notanodised_tilt_1}
    \end{figure}

\begin{figure}[h!]
        \centering
        \includegraphics[width=0.45\textwidth]{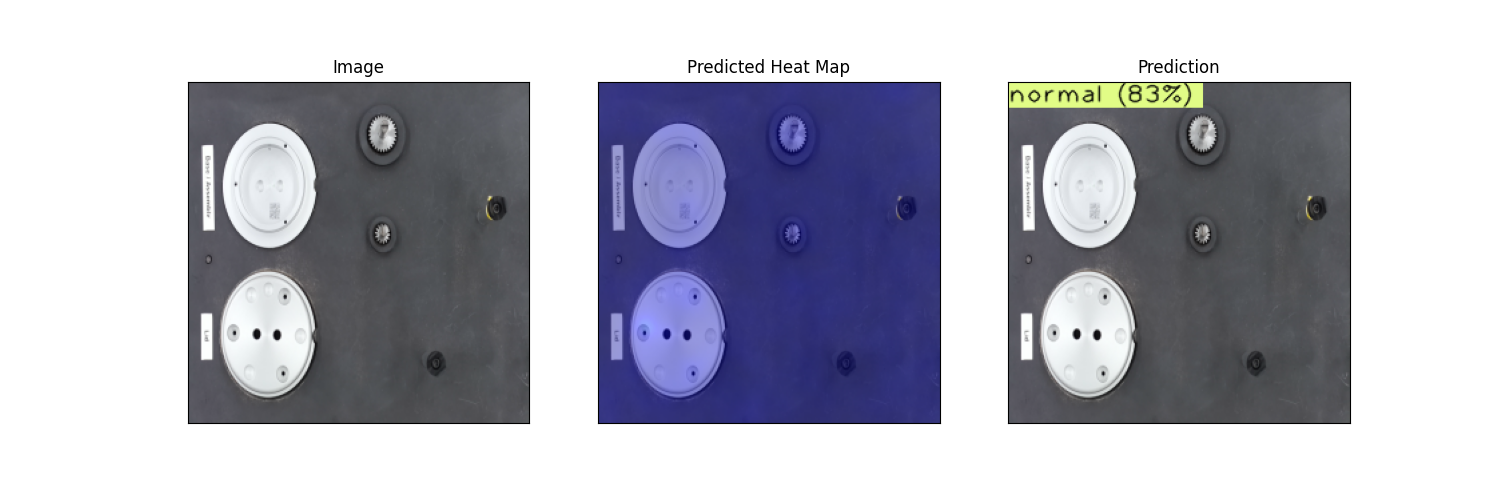}
        \caption{Aonomaly Detection Results for Normal Parts}
        \label{fig:normal_inference}
    \end{figure}

\begin{figure}[h!]
        \centering
        \includegraphics[width=0.45\textwidth]{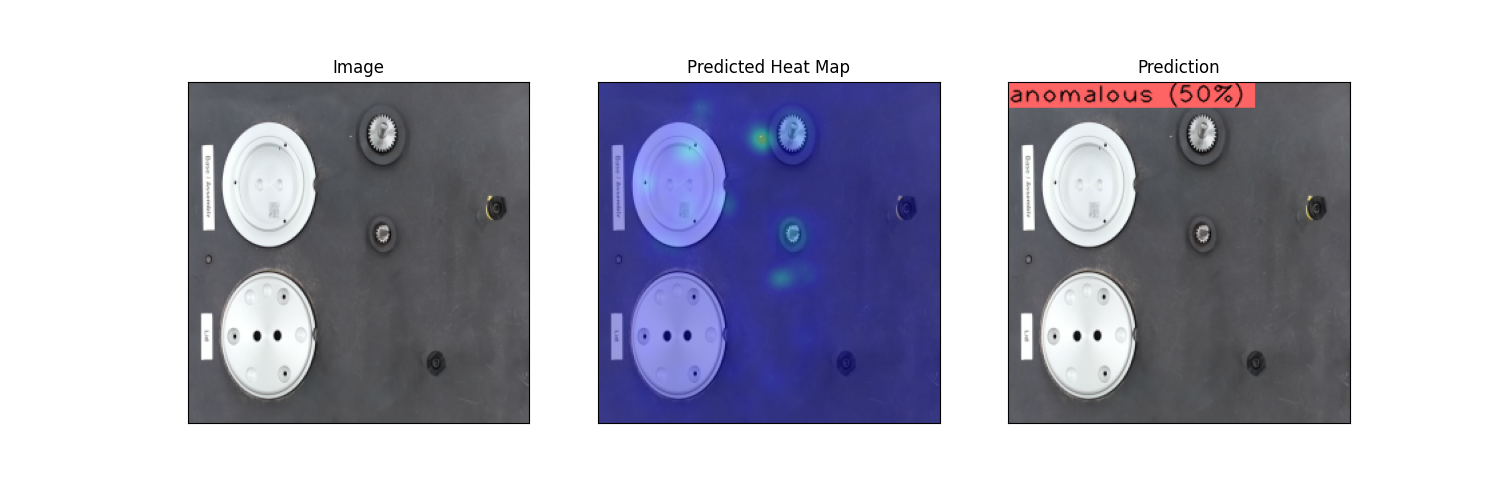}
        \caption{Misclassification in Anomaly Detection for Normal Parts}
        \label{fig:uncertainty}
    \end{figure}

\begin{figure}[h!]
        \centering
        \includegraphics[width=0.45\textwidth]{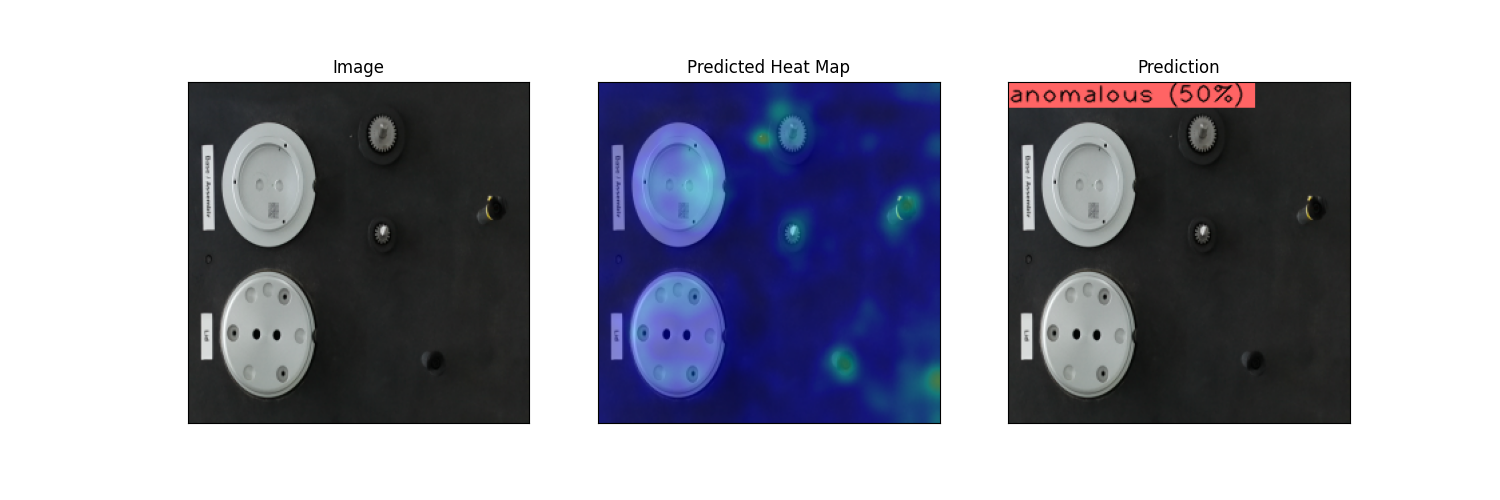}
        \caption{Aonomaly Detection Results for Gears with Missing Teeth}
        \label{fig:teeth-missing}
    \end{figure}

Table \ref{tab:F1 ande AUROC} shows that the model achieves a perfect 100\% score for both F1 Macro and AUROC across all product and setup conditions, with AUROC values provided in parentheses. Additionally, Table \ref{tab:Number of Anomalies} details the number of anomalies detected for each product and setup condition, where each test set consists of 10 images—5 anomalous and 5 normal. These tables demonstrate that the PatchCore algorithm effectively distinguishes between normal and anomalous test images without needing prior labeling of anomalous cases for training. However, it's also important to identify the nature of the anomalies by analyzing the anomalous regions in the images and the associated confidence scores. To enhance the understanding of detected anomalies, Figure \ref{fig:notanodised_tilt_1} provides a detailed explanation of how anomalies in non-anodized casings and tilted gears are identified. The process starts by inputting the test image into the model, which then generates a thermal image that highlights abnormal areas. This thermal image marks anomalies, such as specific regions of the casing and gears, more visible. In addition, confidence scores are provided for each anomaly detection. For example, an image flagged as anomalous has a confidence score of 59\%, indicating moderate certainty in the detection. In contrast, Figure \ref{fig:normal_inference} shows a normal part with a strong confidence score of 83\%. These scores, along with the visual localization of anomalies, offer valuable insights for the operator.

\subsection{Ablation Results}

\begin{table}[H]
\caption{Testing Performance Based on Intel-CPU}
\label{tab:Intel-CPU}
\resizebox{\columnwidth}{!}{%
\begin{tabular}{|llll|}
\hline
\multicolumn{4}{|c|}{CPU Intel i7-10875}                                                                                        \\ \hline
\multicolumn{1}{|l|}{}                   & \multicolumn{1}{l|}{PaDiM(20)}    & \multicolumn{1}{l|}{PaDiM(50)}    & PaDiM(80)    \\ \hline
\multicolumn{1}{|l|}{Test AUROC}         & \multicolumn{1}{l|}{0.59}         & \multicolumn{1}{l|}{0.91}         & 0.94         \\ \hline
\multicolumn{1}{|l|}{Test F1 macro}      & \multicolumn{1}{l|}{0.96}         & \multicolumn{1}{l|}{0.98}         & 0.98         \\ \hline
\multicolumn{1}{|l|}{Training Time (s)}  & \multicolumn{1}{l|}{33.80}        & \multicolumn{1}{l|}{34.80}        & 34.80        \\ \hline
\multicolumn{1}{|l|}{Inference Time (s)} & \multicolumn{1}{l|}{17.10}        & \multicolumn{1}{l|}{17.10}        & 18.20        \\ \hline
\multicolumn{1}{|l|}{} & \multicolumn{1}{l|}{PatchCore(20)} & \multicolumn{1}{l|}{PatchCore(50)} & PatchCore(80) \\ \hline
\multicolumn{1}{|l|}{Test AUROC}         & \multicolumn{1}{l|}{0.96}         & \multicolumn{1}{l|}{0.97}         & 0.98         \\ \hline
\multicolumn{1}{|l|}{Test F1 macro}      & \multicolumn{1}{l|}{0.96}         & \multicolumn{1}{l|}{0.94}         & 0.94         \\ \hline
\multicolumn{1}{|l|}{Training Time (s)}  & \multicolumn{1}{l|}{48.40}        & \multicolumn{1}{l|}{120.60}       & 264.40       \\ \hline
\multicolumn{1}{|l|}{Inference Time (s)} & \multicolumn{1}{l|}{20.40}        & \multicolumn{1}{l|}{21.90}        & 26.40        \\ \hline
\multicolumn{1}{|l|}{}                   & \multicolumn{1}{l|}{CFlow-AD(20)} & \multicolumn{1}{l|}{CFlow-AD(50)} & CFlow-AD(80) \\ \hline
\multicolumn{1}{|l|}{Test AUROC}         & \multicolumn{1}{l|}{0.82}         & \multicolumn{1}{l|}{0.92}         & 0.87         \\ \hline
\multicolumn{1}{|l|}{Test F1 macro}      & \multicolumn{1}{l|}{0.93}         & \multicolumn{1}{l|}{0.94}         & 0.85         \\ \hline
\multicolumn{1}{|l|}{Training Time (s)}  & \multicolumn{1}{l|}{362.10}       & \multicolumn{1}{l|}{910.50}       & 635.40       \\ \hline
\multicolumn{1}{|l|}{Inference Time (s)} & \multicolumn{1}{l|}{34.60}        & \multicolumn{1}{l|}{32.20}        & 41.60        \\ \hline
\multicolumn{1}{|l|}{}                   & \multicolumn{1}{l|}{Fastflow(20)} & \multicolumn{1}{l|}{Fastflow(50)} & Fastflow(80) \\ \hline
\multicolumn{1}{|l|}{Test AUROC}         & \multicolumn{1}{l|}{0.48}         & \multicolumn{1}{l|}{0.64}         & 0.95         \\ \hline
\multicolumn{1}{|l|}{Test F1 macro}      & \multicolumn{1}{l|}{0.94}         & \multicolumn{1}{l|}{0.89}         & 0.93         \\ \hline
\multicolumn{1}{|l|}{Training Time (s)}  & \multicolumn{1}{l|}{120.60}       & \multicolumn{1}{l|}{142.90}       & 739.70       \\ \hline
\multicolumn{1}{|l|}{Inference Time (s)} & \multicolumn{1}{l|}{17.40}        & \multicolumn{1}{l|}{20.40}        & 25.30        \\ \hline
\end{tabular}%
}
\end{table}

\begin{table}[H]
\caption{Testing Performance of the PaDiM Algorithm on Raspberry Pi}
\label{tab:Padim-Raspberry-Pi}
\tiny
\resizebox{\columnwidth}{!}{%
\begin{tabular}{|
>{\columncolor[HTML]{FFFFFF}}l 
>{\columncolor[HTML]{FFFFFF}}l 
>{\columncolor[HTML]{FFFFFF}}l 
>{\columncolor[HTML]{FFFFFF}}l |}
\hline
\multicolumn{4}{|c|}{\cellcolor[HTML]{FFFFFF}{\color[HTML]{000000} Raspberry Pi 4 8GB RAM}} \\ \hline
\multicolumn{1}{|l|}{\cellcolor[HTML]{FFFFFF}{\color[HTML]{000000} \textbf{}}} &
  \multicolumn{1}{c|}{\cellcolor[HTML]{FFFFFF}{\color[HTML]{000000} \textbf{PaDiM(20)}}} &
  \multicolumn{1}{c|}{\cellcolor[HTML]{FFFFFF}{\color[HTML]{000000} \textbf{PaDiM(50)}}} &
  \multicolumn{1}{c|}{\cellcolor[HTML]{FFFFFF}{\color[HTML]{000000} \textbf{PaDiM(80)}}} \\ \hline
\multicolumn{1}{|l|}{\cellcolor[HTML]{FFFFFF}{\color[HTML]{000000} \textbf{Test AUROC}}} &
  \multicolumn{1}{l|}{\cellcolor[HTML]{FFFFFF}{\color[HTML]{000000} 0.954545}} &
  \multicolumn{1}{l|}{\cellcolor[HTML]{FFFFFF}{\color[HTML]{000000} 0.985714}} &
  {\color[HTML]{000000} Failed} \\ \hline
\multicolumn{1}{|l|}{\cellcolor[HTML]{FFFFFF}{\color[HTML]{000000} \textbf{Test F1}}} &
  \multicolumn{1}{l|}{\cellcolor[HTML]{FFFFFF}{\color[HTML]{000000} 0.952381}} &
  \multicolumn{1}{l|}{\cellcolor[HTML]{FFFFFF}{\color[HTML]{000000} 0.933333}} &
  {\color[HTML]{000000} Failed} \\ \hline
\multicolumn{1}{|l|}{\cellcolor[HTML]{FFFFFF}{\color[HTML]{000000} \textbf{Training Time}}} &
  \multicolumn{1}{l|}{\cellcolor[HTML]{FFFFFF}{\color[HTML]{000000} 1min+9s}} &
  \multicolumn{1}{l|}{\cellcolor[HTML]{FFFFFF}{\color[HTML]{000000} 1min+28s}} &
  {\color[HTML]{000000} Failed} \\ \hline
\multicolumn{1}{|l|}{\cellcolor[HTML]{FFFFFF}{\color[HTML]{000000} \textbf{Inference Time}}} &
  \multicolumn{1}{l|}{\cellcolor[HTML]{FFFFFF}{\color[HTML]{000000} 24s}} &
  \multicolumn{1}{l|}{\cellcolor[HTML]{FFFFFF}{\color[HTML]{000000} 29s}} &
  {\color[HTML]{000000} Failed} \\ \hline
\end{tabular}%
}
\end{table}

\begin{table}[H]
\caption{Testing Performance of the Patchcore Algorithm on Raspberry Pi}
\label{tab:Patchcore-Raspberry-Pi}
\tiny
\resizebox{\columnwidth}{!}{%
\begin{tabular}{|
>{\columncolor[HTML]{FFFFFF}}l 
>{\columncolor[HTML]{FFFFFF}}l 
>{\columncolor[HTML]{FFFFFF}}l 
>{\columncolor[HTML]{FFFFFF}}l |}
\hline
\multicolumn{4}{|c|}{\cellcolor[HTML]{FFFFFF}{\color[HTML]{000000} Raspberry Pi 4 8GB RAM}} \\ \hline
\multicolumn{1}{|l|}{\cellcolor[HTML]{FFFFFF}{\color[HTML]{000000} \textbf{}}} &
  \multicolumn{1}{c|}{\cellcolor[HTML]{FFFFFF}{\color[HTML]{000000} \textbf{\begin{tabular}[c]{@{}c@{}}PatchCore\\ (20)\end{tabular}}}} &
  \multicolumn{1}{c|}{\cellcolor[HTML]{FFFFFF}{\color[HTML]{000000} \textbf{\begin{tabular}[c]{@{}c@{}}PatchCore\\ (50)\end{tabular}}}} &
  \multicolumn{1}{c|}{\cellcolor[HTML]{FFFFFF}{\color[HTML]{000000} \textbf{\begin{tabular}[c]{@{}c@{}}PatchCore\\ (80)\end{tabular}}}} \\ \hline
\multicolumn{1}{|l|}{\cellcolor[HTML]{FFFFFF}{\color[HTML]{000000} \textbf{Test AUROC}}} &
  \multicolumn{1}{l|}{\cellcolor[HTML]{FFFFFF}{\color[HTML]{000000} 0.977273}} &
  \multicolumn{1}{l|}{\cellcolor[HTML]{FFFFFF}{\color[HTML]{000000} Crash}} &
  {\color[HTML]{000000} Failed} \\ \hline
\multicolumn{1}{|l|}{\cellcolor[HTML]{FFFFFF}{\color[HTML]{000000} \textbf{Test F1}}} &
  \multicolumn{1}{l|}{\cellcolor[HTML]{FFFFFF}{\color[HTML]{000000} 0.956522}} &
  \multicolumn{1}{l|}{\cellcolor[HTML]{FFFFFF}{\color[HTML]{000000} Crash}} &
  {\color[HTML]{000000} Failed} \\ \hline
\multicolumn{1}{|l|}{\cellcolor[HTML]{FFFFFF}{\color[HTML]{000000} \textbf{Training Time}}} &
  \multicolumn{1}{l|}{\cellcolor[HTML]{FFFFFF}{\color[HTML]{000000} 3min+39s}} &
  \multicolumn{1}{l|}{\cellcolor[HTML]{FFFFFF}{\color[HTML]{000000} Crash}} &
  {\color[HTML]{000000} Failed} \\ \hline
\multicolumn{1}{|l|}{\cellcolor[HTML]{FFFFFF}{\color[HTML]{000000} \textbf{Inference Time}}} &
  \multicolumn{1}{l|}{\cellcolor[HTML]{FFFFFF}{\color[HTML]{000000} 1min+10s}} &
  \multicolumn{1}{l|}{\cellcolor[HTML]{FFFFFF}{\color[HTML]{000000} Crash}} &
  {\color[HTML]{000000} Failed} \\ \hline
\end{tabular}%
}
\end{table}
The ablation study shows the results of running the different models on a Raspberry Pi and a general PC with different quantities of images for training. Since our model is tailored for fraud detection, where missing a fraud case (false negative) is more detrimental than a false alarm (false positive), prioritizing the F1 Score over AUROC is crucial. When running the models on the PC, the PaDiM algorithm, as detailed in Table \ref{tab:Intel-CPU}, consistently exhibits robust and superior performance compared to other models. Notably, PaDiM is the fastest model, with training times of 34.8 seconds and inference times of 18.2 seconds for a dataset with 80 photos. However, its accuracy is somewhat reduced by the smaller dataset size. In contrast, PatchCore, shown in Table \ref{tab:Intel-CPU}, demonstrates slightly lower F1 scores and slower training and inference times but remains highly effective. On the other hand, the table above also reveals that CFlow-AD and Fastflow algorithms are less effective in detecting anomalies on our datasets and consume significantly more time, with CFlow-AD taking over 2000\% longer than PaDiM. Consequently, PaDiM and PatchCore emerge as the primary algorithms for visual anomaly detection due to their stability, overall performance, and training efficiency.

Results are shown for the Raspberry Pi platform with PaDiM, in Table \ref{tab:Padim-Raspberry-Pi}, and PatchCore in Table \ref{tab:Patchcore-Raspberry-Pi}. The Raspberry Pi encountered difficulties as the number of training images increased, leading to system crashes and training failures. CFlow-AD and FastFlow encountered crashes during model training on the Raspberry Pi across all the training set sizes.  Nevertheless, when data volume is kept at 20 images, PaDiM achieves training in approximately 1 minute with a testing F1 score of 0.95, nearly matching its performance on a general PC. This limit of 20 training images could limit the system's use in complex applications where lots of training images are required. These findings underscore the need to balance training speed, performance stability, and efficiency when deploying algorithms like PaDiM and PatchCore on resource-constrained platforms such as the Raspberry Pi. The inference (running) time on the Raspberry Pi is also quite high, with 24 seconds for PaDiM and 70 seconds for PatchCore. This means the proposed system would not work on a high-frequency system when checks need to be performed faster than this time. 

\subsection{System stability}

Although the Raspberry Pi-based anomaly detection system is cost-effective and feasible, evaluating its reliability during model inference is crucial. For example, as shown in Figure \ref{fig:uncertainty}, the model misidentifies a normal part as anomalous under extremely dark conditions with only 50\% confidence. This highlights the model's uncertainty, suggesting that additional checks are needed to improve the system's reliability for industrial applications.

\subsection{Pixel impact}

The detection system performs well at identifying component placement errors and missing parts. However, it shows lower confidence and occasional misalignment when detecting subtle component damage and material changes. For instance, in Figure \ref{fig:teeth-missing}, while the system detects anomalies in the image, it struggles to precisely locate the missing teeth in the gears, particularly when the teeth are slightly blurred. Therefore, it is crucial to train the system using images of varying pixel resolutions and compare the classification results to determine if the low-cost system can accurately detect anomalies at lower pixel levels.

\section{Conclusions} \label{sec: conclusion} 

This research showcases the development of a low-cost visual anomaly detection system using affordable hardware like a Raspberry Pi and its camera.It explores how state-of-the-art algorithms like PatchCore from Anomalib can be applied and accelerated for deep learning training and inference on low-cost hardware. The effectiveness of the system is demonstrated through a case study on detecting anomalies in gearbox parts. The model, trained on normal images using a Raspberry Pi, achieved perfect AUROC and F1 scores, successfully detecting various anomalies. It also localizes defects and provides confidence scores, aiding workers in accurately identifying specific issues during production. Ablation experiments comparing performance on low-cost hardware and general PCs show that unsupervised models like PaDiM and PatchCore perform similarly, with fast inference times. However, the low-cost system is limited by its capacity for training and inference data, and struggles to distinguish between anomalies caused by product defects and setup changes (e.g., lighting or tray movement), potentially hindering its use in fast production lines.

This paper presents a pioneering exploration of developing a low-cost visual anomaly detection system using pre-trained deep learning models in resource-constrained environments. The implementation of deep learning training and inference on a Raspberry Pi with low-cost cameras faces challenges due to the limited image resolution and the large volume of data, which poses difficulties for the practical application of such cost-effective solutions in industrial production. Future research should explore model distillation or data compression to improve the deployment of deep learning models on edge devices, enhancing the reliability and stability of low-cost visual detection systems.

{\small
 	\bibliographystyle{chicago}
 	\bibliography{reference.bib}
}

\end{document}